\pdfoutput=1

\documentclass[11pt]{article}

\usepackage[final]{acl}
\usepackage[T1]{fontenc}
\usepackage{multirow}
\usepackage{multicol}
\usepackage[english]{babel}

\usepackage{floatrow}
\newfloatcommand{capbtabbox}{table}[][\FBwidth]
\usepackage{listings}
\usepackage{cmap}					
\usepackage{mathtext} 				
\usepackage[T2A]{fontenc}			
\usepackage[utf8]{inputenc}			

\usepackage{graphicx}
\graphicspath{{images/}}
\usepackage{listings}
\usepackage{tabularx}
\usepackage{array}
\usepackage{makecell}
\usepackage{enumitem}
\usepackage{subcaption}

\begin{document}

\title{StRuCom: A Novel Dataset of Structured Code Comments in Russian}

\author{
 \textbf{Maria Dziuba\textsuperscript{1,2}},
 \textbf{Valentin Malykh\textsuperscript{1,3}},
\\
 \textsuperscript{1}MTS AI,
 \textsuperscript{2}ITMO University,
 \textsuperscript{3}IITU University,
\\
 \texttt{{dziuba.maria}@niuitmo.ru}
 \\
 \texttt{valentin.malykh@phystech.edu}
}

\maketitle   
\begin{abstract}
Structured code comments in \textit{docstring} format are essential for code comprehension and maintenance, but existing machine learning models for their generation perform poorly for Russian compared to English. To bridge this gap, we present StRuCom — the first large-scale dataset (153K examples) specifically designed for Russian code documentation. Unlike machine-translated English datasets that distort terminology (e.g., technical loanwords vs. literal translations) and docstring structures, StRuCom combines human-written comments from Russian GitHub repositories with synthetically generated ones, ensuring compliance with Python, Java, JavaScript, C\#, and Go standards through automated validation. Fine-tuning Qwen2.5-Coder models (0.5B-7B) on StRuCom shows statistically significant improvements of \textit{chrf++} and \textit{BERTScore} over baseline models.

\end{abstract}

\section{Introduction} 

The automated generation of structured code comments in \textit{docstring} format, including detailed descriptions of functionality, parameters, return values, exceptions, and usage examples, greatly improves codebase maintenance. Structured code comments provide developers with quick and easy access to the required information, and can also be used to automatically generate project documentation, for instance, in HTML format. However, modern language models, such as Qwen2.5-Coder \cite{hui2024qwen2} and DeepSeek-Coder \cite{guo2024deepseek}, primarily focus on English-language data and therefore perform poorly for Russian-language comment, neglecting the needs of Russian-speaking developers. These developers, working on localized projects, who often encounter linguistic barriers, which can lead to code misunderstanding and a waste of time. In view of this, there is a strong need for a specialized model for this task, which requires curated training data.

Unfortunately, existing datasets (English-centric CodeSearchNet \cite{Husain2019CodeSearchNetCE} or multilingual MCoNaLa \cite{Wang2022MCoNaLaAB}) mostly focus on code summarization and retrieval tasks, not on function-level documentation generation. The datasets that contain both simple comments and docstrings in English (for example, the Vault \cite{Mnh2023TheVA}), firstly, require a tool for structure-based filtration to check comments for existence of detailed functionality descriptions, covering all function parameters, exceptions and its return value. Secondly, machine translation of English comments cannot be straightforwardly used, as it introduces distortions (e.g., translating ``endpoint'' as ``конечная точка'' instead of the established loanword ``эндпоинт'') \cite{Wang2022MCoNaLaAB} and disrupts \textit{docstring} structure.  

In this work, we present StRuCom, the first specialized dataset for generating structured Russian-language code comments. To create it, we developed a tool for filtering and validating comment structures, supporting five popular documentation styles: Python - GoogleDoc\footnote{\url{https://google.github.io/styleguide/pyguide.html}}, JavaScript - JSDoc\footnote{\url{https://jsdoc.app}}, Java - JavaDoc\footnote{\url{https://docs.oracle.com/javase/8/docs/technotes/tools/windows/javadoc.html}}, C\# - XML\footnote{\url{https://learn.microsoft.com/en-us/dotnet/csharp/language-reference/xmldoc/recommended-tags}}, and Go - GoDoc\footnote{\url{https://tip.golang.org/doc/comment}}. The dataset combines real-world comments from Russian repositories with synthetically generated examples. Using this data, we finetuned the Qwen2.5-Coder model family (0.5B, 1.5B, 3B, and 7B parameters), demonstrating statistically significant improvements in generation quality via \texttt{chrf++} \cite{popovic2017chrf++} and \texttt{BERTScore} \cite{zhangbertscore} metrics compared to baseline versions.

\textbf{Our contributions:}  
\begin{enumerate}
\item \textbf{Filtering tool for structured comments.} We developed an automated tool to validate comment structures across five documentation standards (Python, Java, Go, C\#, JavaScript). 
\item  \textbf{Dataset.} We compiled a dataset of 270,000 Russian-language code-comment pairs, combining real-world examples from GitHub repositories with synthetically generated annotations for five programming languages.  
\item \textbf{Models.} We finetuned Qwen2.5-Coder models (0.5B–7B parameters), achieving statistically significant improvements in comment generation quality (measured by \texttt{chrf++} and \textit{BERTScore}) over base models.  
\end{enumerate}

\section{Related Work}

\subsection{Datasets}

The existing datasets for code-to-text tasks are mainly focused on English-language content. \textbf{The Stack} ~\cite{Kocetkov2022TheS3} combines multilingual code from 658 programming languages (67 TB in version 2.x), collected from a variety of sources: Software Heritage Archive, GitHub Issues, Stack Overflow, etc. Despite its scale, the set is not adapted for supervised fine-tuning (SFT) tasks and requires significant preprocessing. \textbf{The Vault} ~\cite{Mnh2023TheVA}, derived from The Stack v1, includes 43 million English-language code-text pairs from 10 programming languages. The data was obtained by extracting docstrings and inline comments using the \textit{Code-Text} parser \footnote{\url{https://github.com/FSoft-AI4Code/CodeText-parser/tree/main}}. However, structured comments (with parameters and usage examples) remain rare, which is partly explained by the predominance of short functions in the source data. \textbf{CodeSearchNet} ~\cite{Husain2019CodeSearchNetCE}, part of the CodeXGLUE benchmark ~\cite{codexglue}, contains 1 million English-language code-text pairs for 6 languages. The set is focused on code search: text descriptions are limited to the first paragraphs of the documentation, which simplifies comparison, but excludes complex descriptions. \textbf{MCoNaLa} \cite{Wang2022MCoNaLaAB} offers limited multilingual support: 345 Russian, 341 Spanish, and 210 Japanese intent-snippet pairs for Python. The focus on narrow ``how-to'' scenarios and a small size limit the applicability of this dataset for structured documentation tasks. 

\subsection{Models}

The rapid advancement of large language models (LLMs) has enabled breakthroughs in automated code documentation. While proprietary models (e.g., GPT-4 \footnote{\url{https://openai.com/index/gpt-4/}}) have an ability to solve in these tasks, their closed-source nature limits adoption for security-sensitive applications in industry. We focus on open-source code-specialized models under 7B parameters — a practical size for local deployment — analyzing their suitability for multilingual comment generation. Although these models, for instance \textbf{DeepSeek-Coder (1.3B–6.7B)} and \textbf{Qwen2.5-Coder (0.5B–7B)}, offer support for Russian language, they perform poorly for Russian-language documentation generations and their quality has not previously been evaluated due to the absence of benchmarks for this task.

\section{Dataset}

\subsection{Collection Process}

To construct our dataset, we crawled all existing Russian-language repositories on GitHub for the selected programming languages (Python, Java, JavaScript (JS), C\#, and Go). Since the GitHub API does not provide a direct query to identify the natural language used by repository authors, we developed a novel approach to address this limitation. Our program retrieved repositories with Russian-language descriptions and permissive licenses (allowing commercial use or lacking licensing restrictions). The crawled repositories contained comments written in various languages. For details on comment extraction see Appendix \ref{sec:comment_extraction}.

\subsection{Filtration Process}
At the initial stage of filtering, all comments were standardized to follow a uniform style based on the conventions established for each programming language: Python - GoogleDoc, JavaScript - JSDoc, Java - JavaDoc, C\# - XML, and Go - GoDoc. Examples of these standardized formats are provided in Appendix \ref{sec:com_str}. 
To further divide comments into types by structure, we suggest the following terminology: \textbf{A structured comment} is a comment that can be parsed by the \textit{docstring\_parser} library \footnote{\url{https://github.com/nmd2k/docstring\_parser }} and contains either parameter lists, return value descriptions, or exception descriptions. \textbf{A complete comment} is a structured comment that provides a comprehensive description of all its component parts, including types (if needed). \textbf{An incomplete comment} is a structured comment that lacks a description of any of its component parts, which is why it cannot be called complete. \textbf{Unstructured comments} are those that do not correspond to a specific format used in a given programming language. For more information about filtration by structure see Appendix \ref{sec:filtration}. Only structured and complete comments were included in the final version of the dataset.

\subsection{Enhancement with LLM}  
Based on the statistics on the structuredness of the collected data from GitHub, many code comments are incomplete or unstructured and generally of poor quality. For some programming languages (for example, JavaScript and Python), there is very little data and this is not enough to finetune neural networks. To solve these problems, we used large language models (LLM), generating synthetic data using them in two ways: generating comments from scratch and improving existing comments. For additional information about comment's enhancement see Appendix \ref{sec:enhancement}.

\subsection{Dataset Overview}
The Table \ref{tab:final_stats_data} presents the final statistical data of the final set, combining synthetic (improved by the Miqu-70B model and generated from scratch by Qwen2.5-Coder-32B-Instruct) and real comments from more than 150,000 Russian-language GitHub repositories of five programming languages: Python, Java, Go, C\# and JavaScript. The total amount of data is 153,181 examples, of which 79,548 are improved, 65,914 are synthetic, and 7,719 are real comments.

\begin{table}
    \centering
    \resizebox{\textwidth}{!}{ 
    \begin{tabular}{|c|c|c|c|}
    \hline
    Prog. lang. & \makecell{Enhanced} & \makecell{From scratch} & Real \\ 
    \hline
    Python       & 14,625 & 10,078 & 359   \\ 
    \hline
    Java        & 16,283 & 10,536 & 2,619 \\ 
    \hline
    Go         & 7,278  & 20,339 & 232  \\ 
    \hline
    C\#         & 39,715 & 5,617 & 4,435  \\ 
    \hline
    JavaScript  & 1,647 & 19,344 & 100    \\ 
    \hline
    $\sum$      & 79,548 & 65,914 & 7,719 \\ 
    \hline
    \end{tabular} }
\caption{Statistics of the collected Russian-language data on programming languages and methods of obtaining them. The table shows the amount of improved (modification of existing comments by the Miqu-70B model), generated from scratch (synthetic data from Qwen2.5-Coder-32B-Instruct) and real comments.}
\label{tab:final_stats_data}
\end{table}

The uniqueness of the proposed dataset is determined by several factors (see Table \ref{tab:comparison_with_datasets}). Firstly, this is the first large corpus with Russian-language documentation for functions. The only existing dataset with comments in Russian, MCoNaLa, is designed to solve a different problem - searching for a code snippet based on the user's intent and, therefore, is not suitable for generating structured comments in the \textit{docstring} style. Secondly, our dataset was strictly checked for structure and completeness: all comments were modified to one of the formats used in the industry for each specific programming language. In other datasets, either there are no structured comments at all (MCoNaLa, CodeSearchNet), or they have not been filtered by structure (the Vault). Thirdly, as a result of the addition of synthetic data, the proposed set, unlike MCoNaLa, has a sufficient size to train large language models for all five selected programming languages.

\begin{table*}[tbh!]
\centering
\resizebox{\textwidth}{!}{
\begin{tabular}{|c|c|c|c|c|}
\hline
Feature & CSN & Vault & MCoNaLa & Our dataset \\ \hline
\makecell{\#Pairs \\<<code-text>>}& 6.5M & 43K & \makecell{341 - es, 210 - ja, \\ 345 - ru} & 153K \\ \hline
\makecell{Code \\ format} & Functions & \makecell{Functions, classes, snippets} & Code snippets & Functions \\ \hline
\makecell{Text \\ format} & \makecell{Unstr., \\ 1-2 sent.}  & \makecell{Mixed (unstr. and str. w/o \\ filtration by structure)} & \makecell{Unstr., \\ (1-2 sent.)} & \makecell{Str. complete \\ (>5 sent.)} \\ \hline
\makecell{Progr. \\ lang.} & \makecell{Go, Java, PHP, \\  JavaScript,\\ Python, Ruby} & \makecell{Java, JavaScript, Python, \\ Ruby, Rust, Golang,\\ C\#, C++, C, PHP} &  \makecell{Python, Java, \\  JavaScript} & \makecell{Java, Python, C\#,\\ Go, JavaScript} \\ \hline
Nat. lang. & en & en & ru, ja, es & ru \\ \hline
\makecell{Data \\ source} & GitHub & The Stack & Stack Overflow & \makecell{GitHub (Rus. repos.)}\\ 
\hline
\end{tabular}
}
\caption{Comparison of the characteristics of the proposed dataset with existing analogues (CSN, Vault, MCoNaLa) by key parameters. The table shows the amount of data, the formats of code and text representation, the coverage of programming languages, linguistic features and data sources. The dataset we propose stands out with a strict focus on Russian-language structured comments on functions (273 thousand pairs), which contrasts with English-language counterparts operating with unstructured or mixed comments.}
\label{tab:comparison_with_datasets}
\end{table*}

\section{Experiments}
We conducted experiments, where we first benchmark existing open-source code-specific LLMs of different size (Qwen2.5-Coder (0.5B - 7B) and DeepSeek-Coder (1.3B - 6.7B)), then  finetune Qwen2.5-Coder (0.5B - 7B) on 7,500 comments, sampled from a synthetic part of our dataset and evaluate all models on our test set, 500 comments, sampled from real comments.

\paragraph{Evaluation}  
We evaluated the models using standard natural language generation metrics, including chrf++ ~\cite{popovic2017chrf++} and a modified BERTScore ~\cite{zhangbertscore}. Instead of the traditional BERT ~\cite{kenton2019bert}, we employed E5-Mistral 7B ~\cite{wang2022text,wang2023improving}, which offers superior performance for Russian, outperforming BERT models.

\paragraph{Training and Results}
The additional information about training setup, hyperparameters, etc. is located in Appendix \ref{sec:experiments} and in Table \ref{tab:combined_results}.
Finetuning on the proposed dataset significantly improves the quality of comment generation using the BERTScore metric for all model sizes and most languages. For chrf++, significant improvements are observed in small number of cases. The results confirm that the proposed approach is effective for adapting language models to the task of generating Russian-language comments, especially in terms of semantic correctness (BERTScore).

\section{Conclusion}

In this paper, we have developed a tool for filtering structured comments, collected a dataset of 153 thousand Russian-language code-comment pairs (real and synthetic data for 5 programming languages) and further trained the Qwen2.5-Coder (0.5B–7B) models. Experiments have shown a significant improvement in the quality of comment generation using the chrf++ and BERTScore metrics. We plan to expand the dataset by adding other programming languages, and develop and implement a quality criterion for structured code comments to automatically filter data and therefore improve the quality of the dataset.

\section{Limitations}
The study has several limitations, including a specific commenting style limitation, an imbalanced test dataset, and the assumption that code comments always contain useful information about code functionality, which is not always true. Additionally, code comments from GitHub may be redundant, uninformative, or contain errors, negatively impacting the dataset's quality. 

\bibliography{lit}

\appendix
\section{Comment Extraction}
\label{sec:comment_extraction}
To extract comments, we used the \textit{function\_parser}\footnote{\url{https://github.com/ncoop57/function_parser}} tool for Python, Java, and Go. For JavaScript and C\#, we employed \textit{Code-Text}. The GitHub data collection process consisted of several steps. First, code snippets from Python and JavaScript libraries with very few non-English comments were excluded. The formatting of comments in Java, JavaScript, and C\# was then standardized. In C\#, XML tags such as <summary> were corrected. For Java and JavaScript, redundant whitespaces, line breaks in block comments (delimited by /** and */), and HTML tags were removed.
Next, automatically generated comments in C\# and JavaScript were filtered out. Duplicate comments in the function and docstring columns were eliminated, along with duplicates based on function and docstring independently. The language of each comment was then identified using Lingua \footnote{\url{https://github.com/pemistahl/lingua-py}}. More information about language identification methods that we used is in Appendix ~\ref{sec:lang_id}. If Lingua failed to determine the language, the corresponding comments were excluded from the dataset.
To improve language identification accuracy, Lingua was provided with short descriptions of comments, ensuring tags and identifier names that could degrade identification quality were removed. This process was applied to all programming languages except Go, which has a relatively simple comment structure.

The final dataset, after filtering, is summarized in Table \ref{tab:collection_stats}. The results show that JavaScript and Go are characterized by a similar trend: a high proportion of commented repositories (70.8\% and 55.9\%) and functions (70.2\% and 25.8\%) are combined with a low percentage of Russian-language comments (24.0\% and 16.4\%), which may indicate the predominance of English-language documentation in their ecosystems. On the contrary, Python and C\# show an increased proportion of Russian—language comments (49.2\% and 36.4\%), which is probably due to regional development practices - the active participation of Russian-speaking communities in projects in these languages, where comments are often written in their native language for the local context.

\begin{table*}[tbh!]
\centering
\resizebox{\textwidth}{!}{
    \begin{tabular}{cccccccccc}
        \hline
    \multicolumn{1}{c}{\multirow{3}{*}{\makecell{Programming \\ language}}} & \multicolumn{3}{c}{\#Repositories} & \multicolumn{3}{c}{\#Functions} & \multicolumn{3}{c}{\#Comments}  \\ \cline{2-10} 
    \multicolumn{1}{c}{} & \multicolumn{1}{c}{\makecell{With \\ comments}} & \multicolumn{1}{c}{\makecell{Total}} &  \multicolumn{1}{c}{\%} &  \multicolumn{1}{c}{\makecell{With \\ comments}} & \multicolumn{1}{c}{\makecell{Total}}  & \multicolumn{1}{c}{\%}  & \multicolumn{1}{c}{\makecell{in Russian}} & \multicolumn{1}{c}{\makecell{Total}} & \multicolumn{1}{c}{\makecell{\% in \\ Russian}} \\ \hline
        Python & 18,535 & 64,440 & 28.8\% & 305,187 & 1,627,726 & 18.7\% & 150,255 & 305,187 & 49.2\%\\
        Java & 13,525 & 42,271 & 32.0\% & 409,506 & 2,684,650 & 15.3\% & 98,622 & 409,506 & 24.1\%\\
        Go & 2,592 & 4,639 & 55.9\% & 117,691 & 456,347 & 25.8\% & 19,276 & 117,691 & 16.4\% \\
        C\# & 8,858 & 26,329 & 33.6\% & 291,142 & 596,905 & 48.8\% & 106,058 & 291,142 & 36.4\% \\
        JavaScript & 15,073 & 21,291 & 70.8\% & 129,767 & 184,871 & 70.2\% & 31,084 & 129,767 & 24.0\%\\
        \hline
    \end{tabular}
    }
    \caption{Statistics on data collection from GitHub, including analysis of repositories, functions, and comments on programming languages, grouped into three categories: \textbf{repositories} (the total number of repositories for each programming language, the number of at least one comment, and the percentage of the latter), \textbf{functions} (the total number of functions, the number of functions with comments and their relative proportion) and \textbf{comments} (the total number of comments, the number of Russian-language comments and their percentage).}
\label{tab:collection_stats}  
\end{table*}

\section{Language identification}
\label{sec:lang_id}
We applied two language identification methods to determine the language of the comments: FastText~\cite{joulin2017bag,joulin2016fasttext} and Lingua. FastText uses a bag-of-n-grams approach to capture partial word order information, enabling efficient processing of large datasets on consumer hardware. Its pretrained models can classify text into one of 217 supported languages with high speed and efficiency. Lingua, on the other hand, employs a probabilistic n-gram model combined with rule-based heuristics, focusing on achieving high detection accuracy across 75 supported languages.
While FastText offers broad language coverage and high efficiency, it demonstrated high precision but low recall for identifying Russian comments, frequently misclassifying them as less popular languages. Lingua, although slower and more memory-intensive, excels at handling short text and mixed-language inputs, which are common in code comments where natural language often intermixes with programming-specific syntax (e.g., tags and identifier names). Lingua's robustness in these scenarios makes it a preferable choice for detecting natural language within code comments.

\section{Comment Structure}
\label{sec:com_str}
The examples of comment structure for five selected programming languages are shown in Figure \ref{fig:docstyles}. Notably, Python’s GoogleDoc and JavaScript’s JSDoc are the only styles among the selected ones that require explicit descriptions of parameter types and return types, reflecting the dynamically-typed nature of these languages. JSDoc shares stylistic similarities with JavaDoc, emphasizing structured documentation. By contrast, C\# utilizes XML for comment formatting, providing a more tag-based approach. GoDoc stands apart with its flexible and descriptive style, as it imposes no strict format requirements, allowing developers to use a nearly free-form commentary approach.

\begin{figure*}[htb!]
\centering
\lstset{
    basicstyle=\small\ttfamily,
    numbers=none,
    columns=fullflexible,
    keepspaces=true,
    showstringspaces=false,
    tabsize=4,
    frame=single,
}

\begin{minipage}[t]{0.48\textwidth}
\begin{lstlisting}
short description

long description 

Args:
    name1 (type1): description1
    name2 (type2): description2

Returns:
    type: description

Raises:
    type: description
\end{lstlisting}
\subcaption{Python Google docstring style}
\end{minipage}
\hfill
\begin{minipage}[t]{0.48\textwidth}
\begin{lstlisting}
/** 
 * short description
 *
 * long description
 *
 * @param name1 description1
 * @param name2 description2
 * @return description
 * @throws type description
*/
\end{lstlisting}
\subcaption{JavaDoc comment style}
\end{minipage}

\vspace{1em}

\begin{minipage}[t]{0.55\textwidth}
\begin{lstlisting}
/// <summary>
/// description
/// </summary>
///
/// <param name="name1">description1</param>
/// <param name="name2">description2</param>
///
/// <returns>description</returns>
///
/// <exception cref="type">description</exception>
\end{lstlisting}
\subcaption{C\# XML comment style}
\end{minipage}
\hfill
\begin{minipage}[t]{0.41\textwidth}
\begin{lstlisting}
/**
 * short description
 *
 * long description
 *
 * @param {type1} name1 - description1
 * @param {type2} name2 - description2
 * @return {type} description
 * @throws {type} description
*/
\end{lstlisting}
\subcaption{JSDOC comment style}
\end{minipage}

\vspace{1em}

\centering
\begin{minipage}[t]{0.3\textwidth}
\begin{lstlisting}
// NameOfFunction description
\end{lstlisting}
\subcaption{GoDoc comment style}
\end{minipage}

\caption{Comparison of documentation styles in different programming languages}
\label{fig:docstyles}
\end{figure*}

\section{Filtration by structure}
\label{sec:filtration}

For filtration-by-structure stage, we utilized the fork of \textit{docstring\_parser} library \footnote{\url{https://github.com/rr-/docstring_parser}} and \textit{javalang} \footnote{\url{https://github.com/c2nes/javalang}} tools to extract information about comment structure and \textit{Code-Text} to gather information about code structure. We also added missing types in Python comments where possible using \textit{Code-Text}. 
The dataset's collection showed significant differences in structured comments' availability and completeness across programming languages, as summarized in Table ~\ref{tab:structure_stats}. 
The results demonstrate an inverse relationship between the complexity of the commenting standard and the proportion of complete structured comments. Go, with minimal requirements (only the function name at the beginning of the comment), shows the maximum percentage of full comments (56.4\%, 10,880). On the contrary, Python and JavaScript, where standards require specifying types and complex annotations, have an extremely low proportion of complete comments (1.5\% and 1.4\%), with unstructured ones dominating (94,968 and 14,091). Java and C++ with moderately complex standards occupy an intermediate position: 29.8\% and 22.7\% of full comments, respectively, but a significant number of unstructured (48,347 and 30,188). The table confirms that the simpler the syntax of a structured comment, the higher the proportion of its compliance. The extremely high Go score is explained by the simplified standard, and the low Python/JavaScript values are due to the excessive complexity of the requirements, which leads to a preference for unstructured comments.

\begin{table*}[tbh!]
\centering
\resizebox{\textwidth}{!}{
    \begin{tabular}{ccccc}
        \hline
        \multirow{3}{*}{\makecell[cc]{ Programming \\ language}} & \multicolumn{3}{c}{Structured} & \multirow{3}{*}{Non-structured} \\
        \cline{2-4}
        & \makecell{\% complete out \\ of all Russian} & Complete & Incomplete & \\
        \hline
        Python & 1.5\% & 2,176 & 30,115 & 94,968 \\ 
        Java & 29.8\% & 29,367 & 12,221 & 48,347 \\ 
        Go & 56.4\% & 10,880 & - & 8,396 \\
        C\# & 22.7\% & 24,017 & 41,898 & 30,188 \\
        JavaScript & 1.4\% & 431 & 1,484 & 14,091 \\
        \hline
    \end{tabular}
}
\caption{The structure of Russian-language comments on programming languages. For each language, the following are indicated: the percentage of complete structured comments out of the total number of Russian-language comments (\% of the total number), the absolute values of complete and incomplete structured comments, as well as the number of unstructured ones. In Go, the dash in the ``Incomplete'' column is due to a feature of the commenting standard: comments are considered complete if they begin with the function name, which excludes the ``incomplete'' category.}  
\label{tab:structure_stats}
\end{table*}

\section{Enhancement of comments via LLM}
\label{sec:enhancement}
\begin{figure*}
    \centering
    \includegraphics[width=1\linewidth]{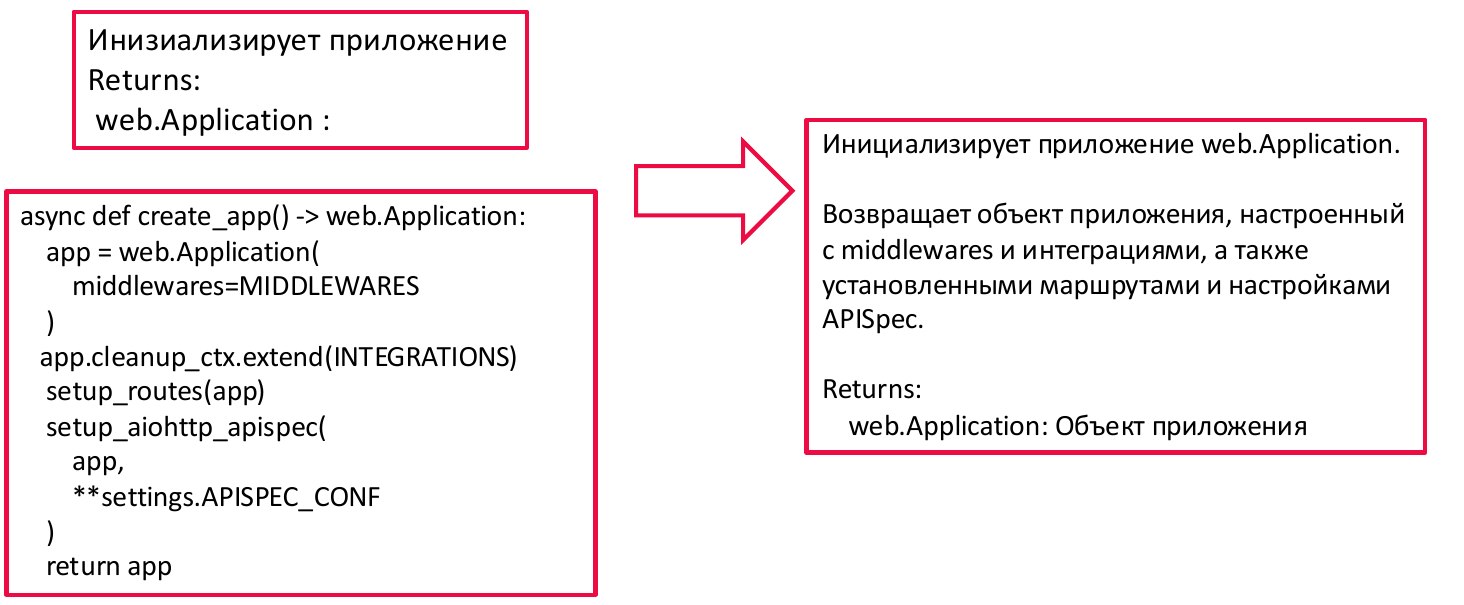}
    \caption{An example of improving a comment. On the left is a function and a comment on it before improvement, which, firstly, has a typo, and secondly, contains a minimum of information about the code. The comment after the improvement is devoid of these shortcomings.}
    \label{fig:function_enhancement_picture}
\end{figure*}

The final dataset includes only those data with the length of both the code and the comment ranging from $250$ to $1,000$ characters. Very short comments and functions were excluded, as the goal was to create a dataset with detailed and comprehensive documentation. Very long comments or features are outliers and therefore were not considered.
Comments were generated from scratch using the Qwen2.5-Coder-32B-Instruct model for functions without comments (see Table \ref{tab:collection_stats}) and for functions, which comments were not successfully enhanced. To improve the dataset, the MIQU 70B \footnote{\url{https://huggingface.co/miqudev/miqu-1-70b }} model was used, which was further trained in Russian. The goal of the improvement is to generate a complete and detailed comment of the best quality based on the function and the existing comment on it. An example is illustrated in figure \ref{fig:function_enhancement_picture}. Candidates for improvement were selected from all the structuredness groups that were not included in the dataset in the ``real'' group. Comment is considered improved if it has become complete as a result of the improvement. Table \ref{tab:enhancement_stats} shows statistics on improving the dataset. Go stands out for the maximum efficiency of improvements (avg = 84.3\%), especially for complete comments (91.5\%), which is explained by a simple commenting standard, where it is enough to specify the function name. Python and JavaScript show the lowest averages (31.9\% and 33.5\%), which is due to the complexity of their standards, which require specifying data types, which makes automatic modification difficult. C\# and Java occupy an intermediate position: C\# shows a high average percentage of improvements (80.1\%) with a peak in the full comments category (92.4\%), while Java shows moderate results (avg = 48.2\%).

\begin{table*}[tbh!]
\centering
\resizebox{\textwidth}{!}{   
    \begin{tabular}{clcccc}
    \hline
    \makecell[c]{Programming \\ language} &  & Non-structured & Incomplete & Complete &  \\ \hline
    \multirow{2}{*}{Python}   & \#Enhanced comments  & 10 775  & 3 455 & 395  & $\sum$ = 14 625  \\ 
    & \% out of the original quantity & 24.2\% & 23.2\% & 48.1\% & avg = 31.9\% \\ 
    \multirow{2}{*}{Java}  & \#Enhanced comments  & 7 066  & 3 810 & 5 407  & $\sum$ = 16 283  \\ 
    & \% out of the original quantity & 32.0\%  & 57.6\% & 55.1\%  & avg = 48.2\% \\ 
    \multirow{2}{*}{Go}  & \#Enhanced comments  & 3 018 & - & 4 260  & $\sum$ = 7 278 \\ 
    & \% out of the original quantity & 77.1\% & - & 91.5\% & avg = 84.3\% \\ 
    \multirow{2}{*}{C\#} & \#Enhanced comments  & 12 467 & 18 148 & 9 100 & $\sum$ = 39 715 \\ 
    & \% \% out of the original quantity & 74.8\% & 73.1\% & 92.4\% & avg = 80.1\% \\ 
    \multirow{2}{*}{JS} & \#Enhanced comments & 1 386 & 164 & 97 & $\sum$ = 1 647 \\ 
    & \% \% out of the original quantity & 20.4\% & 20.4\% & 59.5\% & avg = 33.5\% \\ \hline
    \end{tabular}
    }
  \caption{Statistics on the improvement of Russian-language comments on programming languages, divided into categories: unstructured, incomplete and complete structured comments. For each language, the absolute number of improved comments, the percentage of improvements relative to the initial number in the category (from the Table \ref{tab:structure_stats}), the total number of improvements ($\sum$) and the average percentage of improvements (avg) are indicated. The dash in the category of incomplete comments for Go reflects their absence in the source data due to the simplified standard for documenting functions.}%
  \label{tab:enhancement_stats}
\end{table*}

\section{Training and Results}
\label{sec:experiments}
The models were trained for 5 epochs with a context length of 2000, a learning rate of 1e-4, and a cosine scheduler with a weight decay of 0.1 and a warmup ratio of 0.01. We used LORA \cite{hu2021lora} adapters with a rank of 8, alpha of 16, and a dropout rate of 0.05 for finetuning. From the synthetic part of the dataset, we sampled 1,500 examples for each programming language, resulting in 7,500 examples. For calculating metrics on real data, we sampled 100 examples for each programming language. The comparison is made with the base models to determine the extent to which training on our synthetic dataset improves the quality. Notably, with a batch size of 1, the model takes approximately 20 hours to train on 5 programming languages using DeepSpeed Zero2 \cite{rasley2020deepspeed} on a single A100 GPU. The results are shown in Table \ref{tab:combined_results}.

\begin{table*}[tbh!]
\resizebox{\textwidth}{!}{  
    \centering
    \scriptsize
    \begin{tabular}{lc@{\hspace{1mm}}cc@{\hspace{1mm}}cc@{\hspace{1mm}}cc@{\hspace{1mm}}cc@{\hspace{1mm}}c}
        \hline
        \multirow{2}{*}{Model} & \multicolumn{2}{c}{Python} & \multicolumn{2}{c}{Java} & \multicolumn{2}{c}{Go} & \multicolumn{2}{c}{C\#} & \multicolumn{2}{c}{JavaScript} \\
        \cline{2-11}
        & BERTScore & chrf++ & BERTScore & chrf++ & BERTScore & chrf++ & BERTScore & chrf++ & BERTScore & chrf++ \\
        \hline
        \multicolumn{11}{l}{Baselines} \\
        \hline
        DeepSeek-Coder 1.3B & 0.837 & 18.3 & 0.827 & 19.2 & 0.811 & 10.4 & 0.812 & 18.4 & 0.839 & 24.7 \\
        & ±0.041 & ±9.8 & ±0.040 & ±7.2 & ±0.042 & ±4.5 & ±0.044 & ±16.9 & ±0.038 & ±8.7 \\
        \hline
        DeepSeek-Coder 6.7B & 0.878 & 34.1 & 0.873 & 36.9 & 0.838 & 21.0 & 0.844 & 36.3 & 0.876 & 38.4 \\
        & ±0.043 & ±10.5 & ±0.044 & ±14.2 & ±0.047 & ±11.1 & ±0.052 & ±18.2 & ±0.033 & ±10.9 \\
        \hline
        Qwen2.5-Coder 0.5B & 0.863 & 26.6 & 0.839 & 20.7 & 0.816 & 10.9 & 0.815 & 14.1 & 0.799 & 9.6 \\
        & ±0.052 & ±9.8 & ±0.056 & ±9.3 & ±0.052 & ±5.6 & ±0.052 & ±8.5 & ±0.035 & ±6.1 \\
        \hline
        Qwen2.5-Coder 1.5B & 0.841 & 22.8 & 0.838 & 21.2 & 0.815 & 11.5 & 0.821 & 31.5 & 0.841 & 23.8 \\
        & ±0.045 & ±10.8 & ±0.045 & ±10.5 & ±0.039 & ±5.0 & ±0.051 & ±14.9 & ±0.035 & ±7.9 \\
        \hline
        Qwen2.5-Coder 3B & 0.784 & 14.2 & 0.829 & 17.2 & 0.819 & 11.0 & 0.817 & 25.7 & 0.841 & 23.7 \\
        & ±0.061 & ±8.4 & ±0.039 & ±6.0 & ±0.041 & ±4.4 & ±0.046 & ±15.5 & ±0.033 & ±6.2 \\
        \hline
        Qwen2.5-Coder 7B & 0.880 & 34.3 & 0.873 & 35.0 & 0.854 & 23.5 & 0.847 & 24.3 & 0.872 & 33.5 \\
        & ±0.040 & ±7.7 & ±0.039 & ±9.8 & ±0.039 & ±9.1 & ±0.037 & ±12.2 & ±0.031 & ±7.9 \\
        \hline
        \multicolumn{11}{l}{Finetuned Models} \\
        \hline
        Qwen2.5-Coder 0.5B & 0.873 & \textbf{35.3} & \textbf{0.872} & 39.7 & \textbf{0.859} & 28.7 & \textbf{0.849} & 44.4 & \textbf{0.871} & 40.3 \\
        & ±0.042 & \textbf{±9.0} & \textbf{±0.040} & ±9.8 & \textbf{±0.038} & ±6.8 & \textbf{±0.041} & ±10.2 & \textbf{±0.035} & ±0.03 \\
        \hline
        Qwen2.5-Coder 1.5B & \textbf{0.877} & 34.4 & \textbf{0.880} & 41.6 & \textbf{0.863} & 32.1 & \textbf{0.857} & 45.7 & \textbf{0.877} & 40.3 \\
        & \textbf{±0.040} & ±7.5 & \textbf{±0.036} & ±8.8 & \textbf{±0.035} & ±6.3 & \textbf{±0.038} & ±9.3 & \textbf{±0.031} & ±0.03 \\
        \hline
        Qwen2.5-Coder 3B & \textbf{0.880} & 34.9 & \textbf{0.881} & 40.6 & \textbf{0.864} & 32.5 & \textbf{0.859} & 46.4 & \textbf{0.878} & 41.3 \\
        & \textbf{±0.040} & ±7.5 & \textbf{±0.035} & ±8.3 & \textbf{±0.035} & ±6.2 & \textbf{±0.037} & ±9.7 & \textbf{±0.031} & ±8.5 \\
        \hline
        Qwen2.5-Coder 7B & 0.878 & 35.5 & 0.882 & \textbf{42.0} & \textbf{0.867} & 32.9 & \textbf{0.859} & 45.9 & 0.879 & \textbf{41.4} \\
        & ±0.039 & ±7.3 & ±0.036 & \textbf{±8.9} & \textbf{±0.035} & ±6.2 & \textbf{±0.034} & ±9.5 & ±0.032 & \textbf{±7.6} \\
        \hline
    \end{tabular}
}
\caption{Comparison of base and finetuned models using BERTScore and chrf++ metrics with statistical significance testing (Mann-Whitney criterion). Statistically significant improvements (p < 0.05) are highlighted in \textbf{bold} when comparing the finetuned model with the corresponding sized base version. The values are presented as the average ± standard deviation.}
\label{tab:combined_results}
\end{table*}

\end{document}